%% file: main.tex
\documentclass[journal]{IEEEtran}
\usepackage{graphicx}
\usepackage{algorithm}
\usepackage{tabularx}
\usepackage{multirow}
\usepackage{amssymb}
\ifCLASSINFOpdf
\else
\fi
\usepackage{amsmath}
\usepackage{xcolor}
\RequirePackage{xcolor, colortbl}
\usepackage{algorithmic}
\usepackage{url}
\usepackage{orcidlink}

\begin{document}
%
\title{CLFT: Camera-LiDAR Fusion Transformer for Semantic Segmentation in Autonomous Driving}
%
%
%

\author{Junyi Gu \orcidlink{0000-0002-5976-6698}, 
        Mauro~Bellone \orcidlink{0000-0003-3692-0688},
        Tom\'a\v{s} Pivo\v{n}ka \orcidlink{0000-0002-1466-5156}, 
        and~Raivo~Sell \orcidlink{0000-0003-1409-0206}~
\thanks{Corresponding Author: Mauro Bellone}
\thanks{J. Gu (junyi.gu@taltech.ee) and R. Sell(raivo.sell@taltech.ee) are with the Department of Mechanical and Industrial Engineering, Tallinn University of Technology, Estonia.}
\thanks{M. Bellone (mauro.bellone@taltech.ee) is with FinEst Centre for Smart Cities, Tallinn University of Technology, Estonia.}
\thanks{Tom\'a\v{s} Pivo\v{n}ka (tomas.pivonka@cvut.cz) is with Czech Institute of Informatics, Robotics, and Cybernetics and Department of Cybernetics, Czech Technical University in Prague, Czech Republic.}}

%
%

\markboth{IEEE Transactions on Intelligent Vehicles, August~2024}%
{Shell \MakeLowercase{\textit{et al.}}: Bare Demo of IEEEtran.cls for IEEE Journals}
%



\maketitle

\begin{abstract}
Critical research about camera-and-LiDAR-based semantic object segmentation for autonomous driving significantly benefited from the recent development of deep learning. 
Specifically, the vision transformer is the novel ground-breaker that successfully brought the multi-head-attention mechanism to computer vision applications. 
Therefore, we propose a vision-transformer-based network to carry out camera-LiDAR fusion for semantic segmentation applied to autonomous driving. 
Our proposal uses the novel progressive-assemble strategy of vision transformers on a double-direction network and then integrates the results in a cross-fusion strategy over the transformer decoder layers. 
Unlike other works in the literature, our camera-LiDAR fusion transformers have been evaluated in challenging conditions like rain and low illumination, showing robust performance. 
The paper reports the segmentation results over the vehicle and human classes in different modalities: camera-only, LiDAR-only, and camera-LiDAR fusion. 
We perform coherent controlled benchmark experiments of the camera-LiDAR fusion transformer (CLFT) against other networks that are also designed for semantic segmentation. 
The experiments aim to evaluate the performance of CLFT independently from two perspectives: multimodal sensor fusion and backbone architectures. 
The quantitative assessments show our CLFT networks yield an improvement of up to 10\% for challenging dark-wet conditions 
when comparing with Fully-Convolutional-Neural-Network-based (FCN) camera-LiDAR fusion neural network. 
Contrasting to the network with transformer backbone but using single modality input, the all-around improvement is 5-10\%. 

Our full code is available online for an interactive demonstration and application\footnote{ \url{https://github.com/Claud1234/CLFT}}.
\end{abstract}


\IEEEpeerreviewmaketitle

\section{Introduction}
\label{sec:intro}

Semantic segmentation of the surrounding environment is a challenging topic in autonomous driving and plays a critical role in various intelligent-vehicle-related research-tasks such as maneuvering, path planning \cite{bartolomei2020perception} \cite{9256269}, and scene understanding \cite{6728473}. 
The field of semantic segmentation has greatly advanced due to the evolution of deep neural networks, particularly Convolutional Neural Networks (CNN), along with the availability of open datasets.
Early studies \cite{maskrcnn} took camera RGB images as input and tested them with datasets that had relatively monotonous scenarios \cite{geiger2013vision}. 
In recent years, the blooming of perceptive sensor industries and strict safety requirements motivated semantic segmentation research related to different sensors and comprehensive scenarios. 
LiDAR sensors are involved the most in all kinds of research. 
Examples of the popular LiDAR-only methods include VoxNet \cite{7353481}, PointNet \cite{qi2017pointnet}, and RotationNet \cite{kanezaki2018rotationnet}. 
However, multimodal sensor fusion is perceived as a promising technique to solve the problem of autonomous driving and has become the mainstream option for semantic segmentation \cite{cui2021deep}.

As an applied research, the advancement of semantic segmentation is driven by the proposals of neural network backbones. 
One of the most popular neural networks recently proposed is the transformer \cite{vaswani2017attention}, which implemented the multi-head attention mechanism \cite{bahdanau2015neural} into the Natural Language Processing (NLP) application. 
The proposal of the Vision Transformer (ViT) \cite{dosovitskiy2020image} inspired researchers to explore its potential in environment perception for autonomous driving. 
In this work, we introduce the camera-LiDAR fusion transformer (CLFT). 
CLFT maintains the generic encoder-decoder architecture of a transformer-based network but uses a novel progressive-assemble strategy of vision transformers on a double-direction network. 
The results of the two network directions are then integrated using a cross-fusion strategy over the transformer decoder layers. 

The CLFT aims to address the following issues that are challenging and less explored in the autonomous driving community.

(i) \textbf{Unbalanced sample distribution.} In real-traffic scenarios, dealing with an unbalanced sample distribution poses a significant challenge for autonomous vehicles. 
For instance, while vehicle lanes consistently have more cars than humans (primarily encountered at crossings or sidewalks), achieving precise perception of human entities remains paramount for the optimal functioning of any autonomous vehicle.
Our previous camera-LiDAR FCN-based fusion model (CLFCN) \cite{gu2022object} achieved more than 90\% accuracy in vehicle classification. However, its accuracy in the human class is limited, reaching only 50\%. 
Due to the under-representation of the human class in the dataset, CNNs face challenges in effectively learning knowledge during explicit down-sampling processes.
In contrast, vision transformers maintain a consistent resolution for representations across all stages. 
Furthermore, their incorporation of a multi-head self-attention mechanism inherently provides an advantage in handling global context, making them more adept at addressing challenges associated with imbalanced class distributions.

(ii) \textbf{The consistency of multimodal input data formats.} LiDAR sensors have attracted broad interest from autonomous driving community and there are different strategies to process the LiDAR's point clouds data \cite{feng2020deep}. 
Unlike previous works in this field that integrate a voxel view of the LiDAR with the camera view \cite{chen2022persformer} \cite{li2022unifying}, our work uses the strategy to project the LiDAR point clouds along $XY$, $YZ$, and $XZ$ plane views; thus, the camera and LiDAR inputs are amalgamated into a unified data representation for subsequent operations, encompassing feature extraction, assembly, and fusion.
Although our CLFT models require the pre-processing of LiDAR point clouds such as calibration, filtering, and projection, we have verified that it is possible to carry out all these operations on the fly based on the current hardware specifications on autonomous vehicles \cite{gu2023endtoend} without significant overhead. 
Together with the inference time analysis in Section \ref{sec:result}, it is possible to claim the practical potential applicability of our models.

The niche of our work compared to other state-of-the-art transformer-based multimodal fusion techniques is detailed in Section \ref{sec:LR}. The contribution of this work can be summarily outlined as follows:
\begin{itemize}
    \item We introduce a new network architecture named CLFT, employing an innovative progressive-assemble strategy of vision transformers within a double-direction network.
    \item To the best of our knowledge \cite{zhong2023transformerbased} \cite{THISANKE2023106669}, CLFT is the first open-source transformer-based network that directly uses camera and LiDAR sensory input for object semantic segmentation tasks.
    \item  We divide datasets based on illumination and weather conditions. This approach allows us to compare and highlight the robustness and efficacy of different models in challenging real-world situations.
    \item We prove the advancement and prospect of multimodal transformer-based models in the autonomous driving perception field, especially the segmentation of under-represented traffic objects.
\end{itemize}

The remainder of the paper is as follows. 
Section \ref{sec:LR} reviews the state-of-the-art literature on camera-LiDAR deep fusion and transformer usage in autonomous driving. 
We analyze the gap in current research and explain how our work contributes to the field. 
Section \ref{sec:method} introduces the CLFT architecture details. 
Section \ref{sec:data_config} presents the pre-processing and configurations of the dataset we used in this work. 
Section \ref{sec:result} reports the experiment results and discussion. 
Finally, a conclusion is conducted in Section \ref{sec:conclu}.

\section{Related work}
\label{sec:LR}
Given the scope of this work, we revisit relevant literature on two aspects of semantic object segmentation for autonomous driving. 
The first part reviews the popular camera-LiDAR fusion-based deep learning proposals. 
The second part presents the recent usage of transformers in autonomous driving research.

\begin{figure*}[t]
\centering
\includegraphics[width=0.99\linewidth]{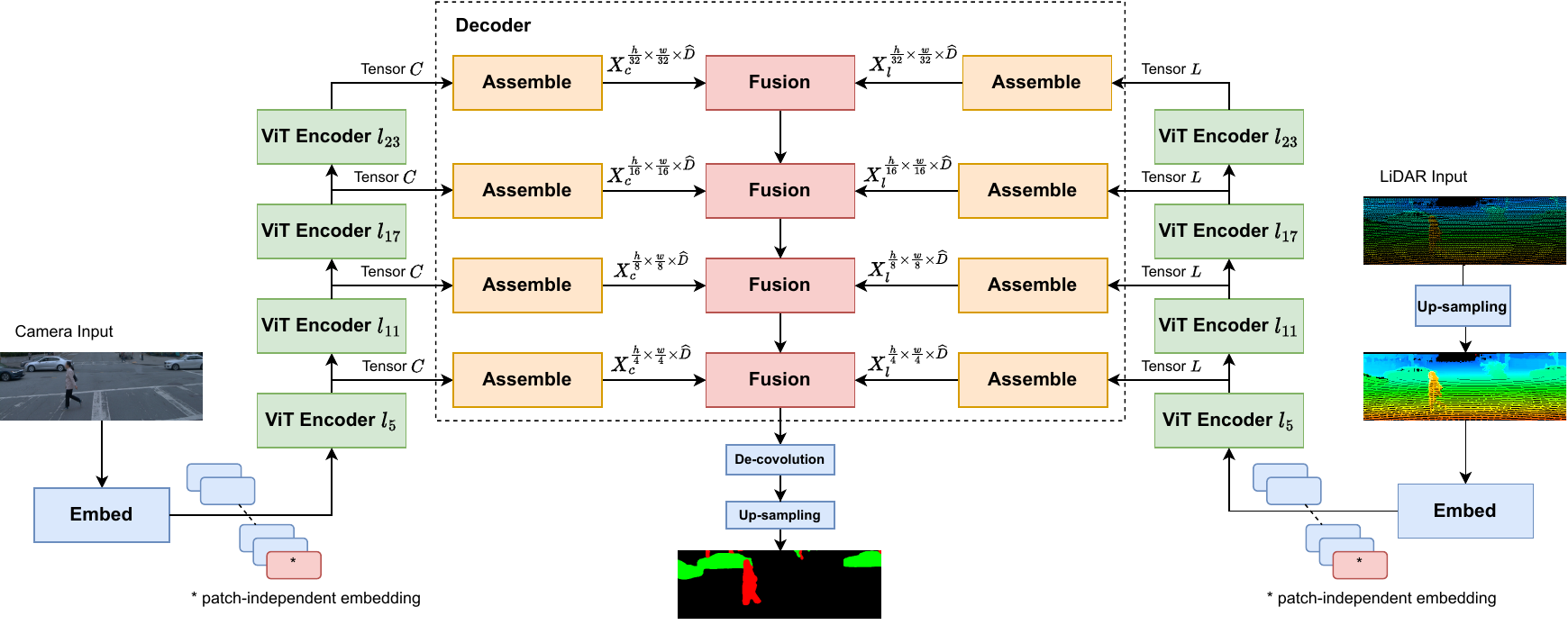}
\caption{The overall architecture of our double-direction network shows camera data flowing from the left side into the ViT encoder, while LiDAR data flows from the right. The camera input is individual RGB channels, and the LiDAR input stands as XY, YZ, and XZ projection planes. The cross-fusion strategy is shown in the center and highlighted using a dashed rectangle.}
\label{fig:overall_architecture}
\end{figure*}

\subsection{Camera-LiDAR fusion-based deep learning}
The fusion of camera and LiDAR data stands out as one of the extensively investigated topics in multimodal fusion, particularly in the context of traffic object detection and segmentation.
Various taxonomies are employed to categorize deep fusion algorithms that integrate camera and LiDAR information.
To distinguish different fusion principles we adopt the patterns suggested in \cite{cui2021deep}, namely \textit{signal-level}, \textit{feature-level}, \textit{result-level}, and \textit{multi-level} fusion. 
This systematic categorization aids in better understanding and comparing the diverse approaches employed in the fusion of camera and LiDAR data for enhanced performance in traffic-related applications.

(i) The \textit{\textbf{signal-level}} fusion is expressed as early-stage fusion as it relies on spatial coordinate matching and raw data (e.g. 2D/3D geometric coordinates, image pixel values) integration to achieve the fusion of two sensing modalities. 
Depth completion \cite{8793637} \cite{cheng2020cspn++} is an iconic application which is instinctively suitable for \textit{signal-level} fusion. 
Work \cite{wulff2018early} \cite{9084150}, and \cite{caltagirone2019lidar} explored the possibility of using \textit{signal-level} fusion in road/lane detection scenarios and its performance-computation trade-off. 
There are relatively few works that implement \textit{signal-level} fusion for traffic object detection and segmentation \cite{dou2019seg} \cite{vora2020pointpainting} because texture information loss is inevitable in sparse mapping and projection process. 

(ii) On the other hand, the literature of \textit{\textbf{feature-level}} fusion is rich. 
In general, the LiDAR data is involved in fusion as either a voxel grid or 2D projection, and the feature map is the most common format for image input.
VoxelNet \cite{zhou2018voxelnet} is the leading work to sample raw point clouds as sparse voxels before the fusion with camera data. 
The examples of the fusion of LiDAR's 2D projections and camera images are \cite{caltagirone2021lidar} \cite{liang2022bevfusion} \cite{valada2020self}.

(iii) The intuition of \textit{\textbf{result-level}} fusion is using the weight-based logical operations to combine the prediction results from different modalities, which is adopted in work \cite{jaritz2020xmuda} \cite{gu20183}. 

(iv) The \textit{\textbf{multi-level}} fusion combines the other three fusion approaches mentioned above to overcome the shortcomings of the respective method. 
Van Gansbeke et al. \cite{8757939} combined \textit{signal-level} and \textit{feature-level} fusion in a network for depth prediction.
PointFusion \cite{xu2018pointfusion} explored the \textit{result-level} and \textit{feature-level} fusion combination by first generating 2D bounding boxes, then filtering the LiDAR points based on these 2D boxes, at last, using a ResNet \cite{he2016deep} and PointNet \cite{qi2017pointnet} network to integrate image and point clouds features to 3D object predictions. 
Other \textit{multi-level} fusion research includes \cite{el2019rgb} \cite{zhao20193d}.

During the literature review, we observe that the transition from \textit{signal/result-level} to \textit{multi-level} fusion is the general trend of camera-LiDAR deep fusion. 
To mitigate some limitations such as computational complexity, early works usually extract geometric information directly from LiDAR data to leverage the existing ready-to-use image processing networks. 
The recent research tends to carry out the fusion in a \textit{multi-level} format, that adopts various fusion strategies and context encoding processes. 
Our work contributes in the line of a \textit{multi-level} fusion architecture which uses a transformer head to encode the input and then execute the cross-fusion of camera and LiDAR data.

\subsection{Transformers in autonomous driving research}

The attention mechanism \cite{bahdanau2015neural} has garnered significant attention from researchers across diverse fields
since its introduction by Vaswani et al. in the transformer architecture for natural language processing (NLP) tasks \cite{vaswani2017attention}. 
Among the most notable transformer variants is the Vision Transformer (ViT) \cite{dosovitskiy2020image}, showcasing its capabilities in computer vision with direct applications in autonomous driving.
Specifically, the autonomous driving perception tasks benefit the most from the attention-mechanism's strengths in global context and long-range dependencies handling. 
In this section, we review the state-of-the-art transformer-based works for 2D and 3D general perception in autonomous driving. 

The 2D perception applications of autonomous driving extract the information from camera images. 
Lane detection is the most prevalent task among 2D perception research. 
Peng et al. \cite{peng2023bevsegformer} proposed a bird's eye view transformer-based architecture for road surface segmentation. 
Work \cite{liu2021end} adopted a lightweight transformer structure for lane shape prediction, first modeled lane markings as regressive polynomials, then optimized the polynomial parameters by a transformer query and Hungarian fitting loss algorithms. 
Other transformer deep networks for road/lane segmentation include \cite{chen2022persformer} \cite{bai2023curveformer}. 
There are relatively fewer works of 2D segmentation because the multimodal fusion is the trend for semantic segmentation in recent. 
Panoptic SegFormer \cite{li2022panoptic} proposed a panoptic segmentation framework utilizing a supervised mask decoder and a query decoupling method to execute the semantic and instance segmentation. 

The research of transformer-based 3D object detection and segmentation is abundant. 
DETR3D \cite{wang2022detr3d} is a variant of the popular DETR \cite{carion2020end} model but extended its 2D object detection potential to 3D detection scenarios. 
DETR3D relied on multi-view images to recover 3D information and used backward geometric projection to combine 2D feature extraction and 3D prediction. 
FUTR3D \cite{chen2023futr3d} is a counterpart network to DETR3D, featuring a modality-agnostic feature sampler designed to accommodate multimodal sensory input for precise 3D bounding box predictions.
PETR \cite{liu2022petr} embedded 3D coordinate information into image to produce 3D position-aware features. 
BEVFormer \cite{li2022bevformer} employed spatial and temporal attention layers for bird’s eye view features to improve the performance of 3D object detection and map segmentation. 
Work \cite{huang2023tri} and \cite{li2023voxformer} focused on the 3D segmentation. 
TPVFormer \cite{huang2023tri} reduced the computational requirement by transforming the volume to three bird’s eye view planes. 
VoxFormer \cite{li2023voxformer} generated 3D voxels from 2D images, then performed cross and self attention mechanisms to 3D voxel queries to compute semantic segmentation results. 

With reference to our review, there are relatively few research works on the semantic object segmentation, let alone the multimodal fusion of camera and LiDAR sensors. 
Work \cite{chen2023futr3d} and \cite{li2022unifying} directly used LiDAR input, but their focus are 3D detection and occupancy prediction. 
Moreover, other latest works \cite{huang2023tri} and \cite{li2023voxformer} produced the voxel and pseudo-point-clouds from the camera input, then carried out the semantic occupancy prediction. 
While our CLFT models directly take LiDAR data as input, and adopt another strategy to process the LiDAR point clouds as image views in camera plane to achieve 2D semantic object segmentation. 
Foremost, our work plays a crucial role in bridging the gap in multimodal semantic object segmentation within the realm of autonomous driving research.

\section{Methodology}
\label{sec:method}
There are two aims of our CLFT models in this work; first is to outperform the existing state-of-the-art single modality transformer-based models; second is to compete with the recent CNN-based models in terms of traffic object segmentation by fusing the camera and LiDAR data.
We maintain the overall structure of the transformer network for dense prediction (DPT) \cite{ranftl2021vision} but invoke a late fusion strategy in its convolutional decoder, which first assemble the LiDAR and camera data in parallel and then integrate their feature map representations. 
We explore the capability of transformer-based networks in semantic segmentation with the advantages of LiDAR sensors, prove transformer networks' potential to classify the less represented samples in contrast with CNNs, at last, provide a late fusion strategy for transformer-related sensor fusion research. 

The encoder-decoder structure has been widely implemented in image analysis transformers. 
We closely follow the protocol of ViT \cite{dosovitskiy2020image} to establish the encoders in our network to create the multi-layer perceptron (MLP) heads for camera and LiDAR data separately.
For the decoders, we refer, but leverage proposals in work \cite{ranftl2021vision} to assemble and integrate the feature representations from camera and LiDAR sensors to create the object segmentation that is more precise than single modality. 
Figure \ref{fig:overall_architecture} shows the overall architecture of our network.

\begin{figure*}
\centering
\includegraphics{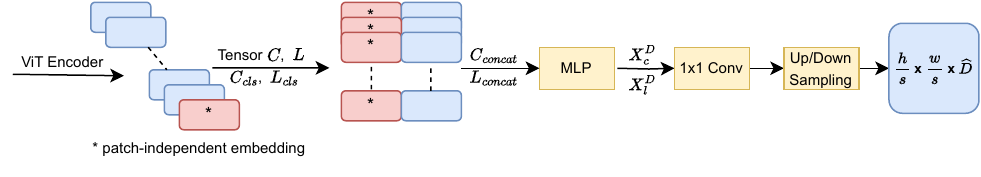}
\caption{Assemble architecture for each transformer decoder block, tokens of each layers are assembled to image-like representations of feature maps. }
\label{fig:assemble_architecture}
\end{figure*}

\begin{figure}
\centering
\includegraphics[width=1.0\linewidth]{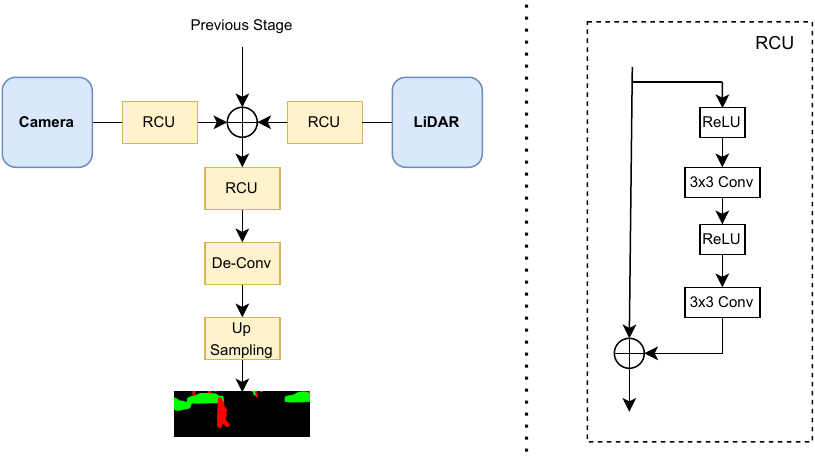}
\caption{Each fusion block receives data from the previous stage and integrates camera-LiDAR data coming from the ViT encoder. Each of this block has residual units, de-convolution, and up-sampling.}
\label{fig:fusion_architecture}
\end{figure}

\subsubsection{Encoder}
ViT innovatively proposed an encoder to convert an image into multiple tokens that can be treated in the same way as words in a sentence; 
consequently, transferred the standard transformer from NLP to computer vision applications.
The ViT encoder uses two different procedures to transfer the images into tokens. 
The first approach divides an image into fixed-size non-overlapping patches, followed by linear projection of their flattened vector representations.
The second approach extracts feature patches from a CNN feature map and then feeds them into the transformer as tokens. 
We retain the ViT's conventions to define the encoder variants in our work, namely, `CLFT-base', `CLFT-large', `CLFT-huge', and `CLFT-hybrid'. 
The `base', `large', and `huge' indicate the encoder's configuration such as layer, size, and amount of parameters. The `hybrid' means other neural network backbones are integrated in the model.
The `CLFT-base', `CLFT-large', and `CLFT-huge' architectures use patch-based embedding methods, have 12, 24, and 32 transformer layers, and the feature dimension $D$ of each token are 768, 1024, and 1280, respectively. 
The `CLFT-hybrid' encoder employs a ResNet50 network to extract pixel features as image embeddings, followed by 12 transformer layers. 
The patch size $p$ of all our experiments is 16. 
The resolution of the input camera and LiDAR image $(h, w)$ is (384, 384), which means the total amount of pixels for each patch $\frac{h * w}{p^2} = 576$ is smaller than feature dimensions $D$ of all variants; thus, the knowledge can be retrieved from input in pixel-wise. 
For the `CLFT-hybrid' encoder, it extracts the features from the input patch of $384 \div 16 = 24$ resolution. 
All the encoders are pretrained using ImageNet \cite{deng2009imagenet}. 
Following work in ViT, we concatenate position embeddings with image embeddings to retain positional information.
Moreover, there is an individual learnable token in sequence for classification purposes. 
This classification token is represented as red block with the asterisk in Figure \ref{fig:overall_architecture}. It is similar to BERT's `class' token \cite{devlin2018bert}, independent from all image patches and positionally embedded. 
Please refer to the original work \cite{dosovitskiy2020image} for the details of these encoder architectures. 


\subsubsection{Decoder}
The transformer networks designed for computer vision usually modify the decoder by implementing convolutional layers at different stages. 
Ranftl et al. \cite{ranftl2021vision} proposed a transformer network for dense prediction (DPT) that progressively assembles tokens from various encoder layers into image-like representations to achieve final dense prediction. 
Inspired by DPT's decoder architecture, we construct a decoder to process the LiDAR and camera tokens in parallel. 

As illustrated in Figure \ref{fig:overall_architecture}, we pick four transformer encoder layers denoted as $t$ ($t=\{2, 5, 8, 11\}$ for `CLFT-base' and `CLFT-hybrid', $t=\{5, 11, 17, 23\}$ for `CLFT-large'), then assemble the tokens from each layer to an image-like representation of feature maps. 
The feature map representations at the initial layers of the network are up-sampled to a high resolution, whereas representations from deep layers ware down-sampled to a low resolution. 
The resolutions are anchored to input image size $(h, w)$, and the sampling coefficients corresponding to encoder layers $t$ are $s=\{4, 8, 16, 32\}$. 
In detail, there are two steps in the assembly process. 
As illustrated in Algorithm \ref{algo:cls_token_projection}, the first step replicates and concatenates the patch-independent `classification token' with all other tokens individually, then forwards the concatenated representations to an MLP process with GELU non-linear activation \cite{hendrycks2016gaussian}. 
The number of individual tokens is denoted as $k$.

 \begin{algorithm}[H]
 \caption{The projection of the `classification token'.}
 \label{algo:cls_token_projection}
 \begin{algorithmic}[1]
 \renewcommand{\algorithmicrequire}{\textbf{Input:}}
 \renewcommand{\algorithmicensure}{\textbf{Output:}}
 \REQUIRE Input tensor $T$, representing either the camera or LiDAR channels containing the `classification token' and patch tokens.
 \ENSURE  Concatenated tensor representations $X_T$
  \STATE $T_{cls} = replicate\{T[:,0]\}$
  \STATE $T_{concat} = T[:,i] \mathbin\Vert T_{cls} \hspace{0.3cm} \forall \hspace{0.2cm} i=1,\dots, k$  
  \STATE $X_T = \hspace{0.1cm} $GELU$(W \cdot T_{concat} + b)$ 
 \end{algorithmic}
 \end{algorithm}

Equation \ref{eq:step_2} shows the second step, which first concatenates the tokens from the first step based on their initial positional order to yield an image-like representation, then passes this representation to two convolution operations. 
The first convolution projects the representation from dimension $D$ to $\hat{D}$ ($\hat{D}$ is set as 256 in our experiments).
The second convolution applies up-sampling and down-sampling toward representation concerning the different layers of transformer encoders. 
$X_c$ and $X_l$ are the concatenated camera and LiDAR representations, $N$ represents the total amount of patches. 
The generic workflows of these two steps are shown in Figure \ref{fig:assemble_architecture}.

\begin{equation}
X_t^{N\times{D}} \Rightarrow X_t^{\frac{h}{s}\times\frac{w}{s}\times\hat{D}} 
\tag{1} \label{eq:step_2}
\end{equation}
\begin{equation}
\notag
X_t = \{ X_c, X_l \} 
\quad
s = \{4, 8, 16, 32 \} 
\end{equation}
\begin{equation}
\notag
t = \{2, 5, 8, 11 \}\ or\ \{5, 11, 17, 23 \}
\end{equation}

The last process of our decoder is the cross-fusion of camera and LiDAR feature maps, which is progressively illustrated in Figure \ref{fig:fusion_architecture}.
We refer to the feature fusion strategy from RefineNet \cite{lin2017refinenet} that forwards the camera and LiDAR representations through two residual convolution units (RCU) in sequence. 
The camera and LiDAR's representations are summed with the results from the previous fusion operation and then went through one additional RCU. 
We pass the output of the last fusion layer to a deconvolutional and up-sampling module to compute the final predicted segmentation. 
The fusion of the information coming from the LiDAR and the camera can happen in any of the fusion block as the connection weights are automatically learned in the network through error back-propagation. 
The  idea of our multiple fusion blocks is to integrate the concept of late-fusion (as each fusion blocks is placed after each assemble block) and the concept of cross-fusion \cite{caltagirone2019lidar} as the connection with each feature map can happen in any of the fusion blocks with different weights. The network automatically learns to weight the best block to integrate tensor information coming from different sensors.

\begin{figure*}[t]
\centering
\includegraphics[width=0.8\linewidth]{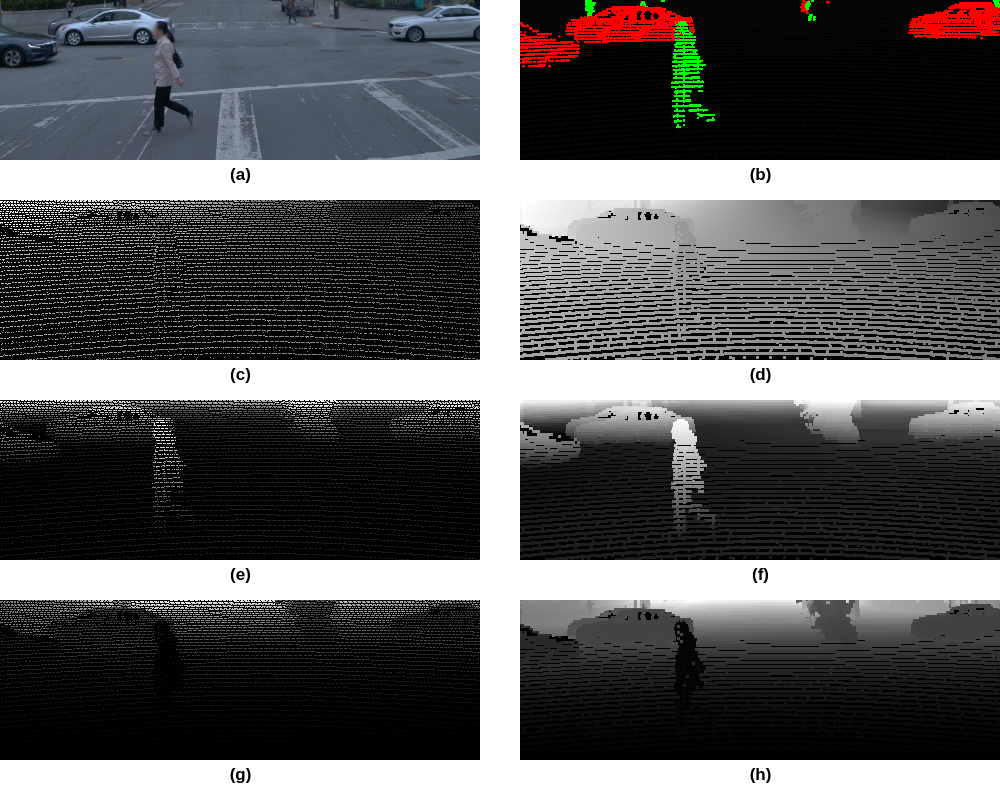}
\caption{Examples of camera image, semantic annotation mask, and pre-processing of LiDAR data. (a) is the RGB image. (b) illustrates the object semantic masks obtained from LiDAR ground truth bounding boxes. (c) (e) (g) are LiDAR projection images in X, Y, Z channels, respectively, while (d) (f) (h) are corresponding up-sampled dense images. Please note that for visualization purposes, the grayscale intensity in (c)-(h) is proportionally scaled based on the numerical 3D coordinate values of the LiDAR points.}
\label{fig: dilated_lidar_anno_imgs}
\end{figure*}

\section{Dataset Configuration}
\label{sec:data_config}
The primary purpose of this work is to compare the performance of the vision transformer and CNN backbones for semantic segmentation. 
Our previous work \cite{gu2022object} successfully modeled and evaluated a ResNet50-based FCN to carry out camera-LiDAR fusion. 
In order to maintain an accordant experiment environment, we construct the input data based on Waymo dataset \cite{sun2020scalability} to evaluate CLFT and other models. 

Waymo dataset is recorded by multiple high-quality cameras and LiDAR sensors. 
The scenes of Waymo dataset span various illumination levels, weather conditions, and traffic scenarios. 
Therefore, as shown in Table \ref{table_datasplit}, we manually partitioned the data sequences into four subsets: light-dry, light-wet, dark-dry, and dark-wet. 
The `light' and `dark' indicate the relative illumination conditions. 
The `dry' and `wet' represent the weather difference in precipitation.

\input{Tables/table_datasplit}

We provide intersection over union (IoU) as the primary indication of model evaluation, with precision and recall values as supplementary information. 
Please note that the IoU is primarily used in object detection applications, in which the output is the bounding box around the object. 
Therefore, We modify the ordinary IoU algorithm to fit the multi-class pixel-wise semantic object segmentation.
The essential change is related to the ambiguous pixels (pixels have no valid labels, details in Section \ref{subsec:object_mask}) that fall out of the class list. 
We assign these pixels as void and exclude them from the evaluation. 
The performance of networks is measured by the statistics of the number of pixels that have identical classes indicated in prediction and ground truth.

\subsection{LiDAR Data Processing}
\label{subsec:lidar_proj}
The LiDAR readings reflect the object's 3D geometric information in the real world. 
Coordinate values in three spatial channels contain features that can be exploited by neural networks. 
As a result, regarding camera-LiDAR fusion, it is common to extract and fuse multi-target features such as images' color textures and point clouds' location information, which is an approach namely as feature-level fusion \cite{8500699}. 

We adopt feature-level fusion in this work. Thus, we project 3D LiDAR point clouds into the camera plane to create 2D occupancy grids in $XY$, $YZ$, and $XZ$ planes.
All the points in LiDAR point clouds are transformed and projected following Equation \ref{eq:2} and \ref{eq:3}, respectively. 
\begin{equation}
\resizebox{0.9\linewidth}{!}{$
    \begin{bmatrix}
        x_t, y_t, z_t
    \end{bmatrix}^T
    =
    \begin{pmatrix}
        r\ p\ y
    \end{pmatrix}
    \begin{pmatrix}
    \begin{bmatrix}
         x_i, y_i, z_i
    \end{bmatrix}^T
    -
    \begin{bmatrix}
         x_c, y_c, z_c
    \end{bmatrix}^T
    \end{pmatrix}
    $}
    \tag{2}\label{eq:2}
\end{equation}
\begin{equation}
\notag
\resizebox{1.\linewidth}{!}{$
    r = 
    \begin{bmatrix}
        1 & 0 & 0 \\
        0 & cos(\rho) & sin(\rho) \\
        0 & -sin(\rho) & cos(\rho) \\
    \end{bmatrix}
    p =
    \begin{bmatrix}
        cos(\theta) & 0 & -sin(\theta) \\
        0 & 1 & 0 \\
        sin(\theta) & 0 & cos(\theta) \\
    \end{bmatrix}
    y =
\begin{bmatrix}
        cos(\phi) & sin(\phi) & 0 \\
        -sin(\phi) & cos(\phi) & 0 \\
        0 & 0 & 1 \\
    \end{bmatrix}
$}
\end{equation}

In Equation \ref{eq:2}, $x_t$, $y_t$, and $z_t$ are the 3D point coordinates after transformation (in camera frame); $r$, $p$, and $y$ represent the Euler rotation matrices to the camera frame with $(\rho, \theta, \phi)$ representing the corresponding Euler angles.
$x_i$, $y_i$, and $z_i$ are the 3D point coordinates before transformation (in LiDAR frame); $x_c$, $y_c$, and $z_c$ denote the camera frame location coordinates.

\begin{equation}
    \begin{pmatrix}
        u, v, 1
    \end{pmatrix}^T
    =
    \begin{bmatrix}
        f_x & 0 & \frac{w}{2} \\
        0 & f_y & \frac{h}{2} \\
        0 & 0 & 1
    \end{bmatrix}
    \begin{pmatrix}
        x, y, z
    \end{pmatrix}^T
   \tag{3}\label{eq:3}
\end{equation}

In Equation \ref{eq:3}, $u$ and $v$ are column and row positions of the point in 2D image plane; $f_x$ and $f_y$ denote camera's horizontal and vertical focal length; $w$ and $h$ represent image resolution; $x$, $y$, and $z$ are transformed 3D point coordinates (same as $x_t$, $y_t$, and $z_t$ in Equation \ref{eq:2}).

 \begin{algorithm}[H]
 \caption{LiDAR points filtering and image pixel values population}
 \label{algo:lidar_point_filtering}
 \begin{algorithmic}[1]
 \renewcommand{\algorithmicrequire}{\textbf{Input:}}
 \renewcommand{\algorithmicensure}{\textbf{Output:}}
 \REQUIRE LiDAR point 3D coordinates $L$, projected LiDAR point coordinates $P$, image resolution $w$ and $h$.
 \ENSURE  LiDAR projection footprints $XY$, $YZ$, and $ZX$.
  \STATE $idx=argwhere(P < \{w,h,+\infty\} \ \&\ P>=\{0,0,0\})$
  \STATE $XY[w\times{h}] \gets 0$
  \STATE $YZ[w\times{h}] \gets 0$
  \STATE $XZ[w\times{h}] \gets 0$
  \STATE $XY[idx] = L[idx,0]$
  \STATE $YZ[idx] = L[idx,1]$
  \STATE $XZ[idx] = L[idx,2]$
 \end{algorithmic}
 \end{algorithm}

\input{Tables/table_fusion}

The operation after transforming and projecting the 3D point clouds into 2D images is filtering, which aims to discard all the points that fall out of the camera view. 
Waymo Open dataset is collected using five LiDAR and five camera sensors covering all vehicle directions. 
This work uses the top LiDAR's point clouds and the front camera's image data. 
As shown in Algorithm \ref{algo:lidar_point_filtering}, three projection footprint images denoted as $XY$, $YZ$, and $ZX$ are generated. The pixels corresponding to 3D points are assigned with 
$x$, $y$, and $z$ coordinates, while the rest are populated with zero. 
At last, we up-sample the LiDAR images before feeding them to machine learning algorithms, as it is a common practice in LiDAR-based object detection research \cite{premebida2014pedestrian} \cite{schlosser2016fusing}. 
Figure \ref{fig: dilated_lidar_anno_imgs} (c)-(g) show the results of the procedure described in this subsection. 

\subsection{Object Semantic Masks}
\label{subsec:object_mask}
Ground truth annotations in Waymo dataset are represented by 2D and 3D bounding boxes, which correspond to camera and LiDAR data separately. 
There are three classes in image annotations: vehicles, pedestrians, and cyclists. Point clouds annotations have an extra class which is traffic signs.
There are two obstacles when using Waymo's ground truth annotations in our networks.

Firstly, vision-transformer-based networks are well-known for requiring vast samples \cite{dosovitskiy2020image}. 
However, the cyclists and traffic signs are relatively rare-represented in the Waymo dataset. 
We notice our CLFT models struggle to learn and predict these two classes in experimental setting as they are less represented in the dataset. 
We assume that with additional data also traffic signs and cyclists can be properly classified. 
Therefore, we discard the traffic signs in this work and merge the cyclists and pedestrians as a new class of so-called human. 

Secondly, our research aims for semantic segmentation, which requires annotations denoted as object contours. 
Since Waymo dataset labeled the object in LiDAR sensor readings as a 3D upright bounding box, we project all the points in the bounding box into the image plane by the same procedure described in Section \ref{subsec:lidar_proj}. 
Figure \ref{fig: dilated_lidar_anno_imgs} (b) shows an example of semantic masks for vehicle and human classes. 
Please note that a limitation of this approach is that some object pixels have no valid labels because there are no corresponding LiDAR points. 

\section{Results}
\label{sec:result}

As mentioned in Section \ref{sec:intro}, our CLFT is the first transformer-based model fusing the camera and LiDAR sensory data for semantic segmentation. 
The experiments in this work focus on the controlled benchmark comparisons in two aspects: i) neural network architecture, ii) input modality.  

The FCN is believed to be the recent generation of deep learning methods with remarkable performance improvements and has become the mainstream for semantic segmentation \cite{mo2022review}. 
Therefore, we choose the CLFCN \cite{gu2022object}, an FCN-based network that fuses camera and LiDAR data for semantic segmentation, as the reference to explore the advantages of transformer backbone. 
Since the transformer is well-known for its strengths in capturing global context and solving long-range dependencies, we expect our transformer-based model to outperform the FCN-based model in scenarios such as unevenly distributed datasets and underrepresented samples. 

Only a few existing deep learning methods process the LiDAR input using the same principle as in this work: representing the 3D point clouds as 2D grid-based feature maps \mbox{\cite{feng2020deep}}.
We compare the CLFT with the Panoptic SegFormer \mbox{\cite{li2022panoptic}} that is also transformer-backbone to evaluate the significance of various input modalities. 
However, the Panoptic SegFormer is purely vision-based. 
We follow the procedures in Section \mbox{\ref{sec:data_config}} to produce the point clouds projection images as LiDAR modality input for Panoptic SegFormer, but the camera-LiDAR-fusion mode is not directly applicable to Panoptic SegFormer.
It is critical to maintain the same input data splits and configurations in experiments for all models. 

\subsection{Experimental setup}
The details of the input dataset configuration are described in Section \ref{sec:data_config}. 
The dataset splits for training, validation, and testing are 60\%, 20\%, and 20\% of the total number of frames, respectively. 
The four data subsets, light-dry, light-wet, dark-dry, and dark-wet, are shuffled and mixed for training and validation but tested individually. 
We adopt the default hyperparameter configurations for CLFCN and Panoptic SegFormer in training. 
Please refer to authors' original work for details \cite{li2022panoptic}.
We employ weighted cross-entropy loss function and Adam optimization \cite{Kingma2014AdamAM} for CLFT networks training. 
The transformer encoder of CLFT is initiated from ImageNet pre-trained weights. 
The transformer decoder and CLFCN's ResNet backbone are initiated randomly. 
The learning rate decay of CLFT networks training follows $l_i=l_0(\alpha^i)$, where $l_0$ is the initial learning rate, and $\alpha$ is 0.99. 
The batch size of CLFT networks training is set as 32 by default, but set as 24 for several experiments that exceed the memory limit, for example, the fusion mode of CLFT-large variant. 
Other hyperparameter settings can be found in the code we public.
The transformer-based networks are trained using an NVIDIA A100 80GB GPU due to the large memory requirement of transformer networks. 
Relatively low-memory-required FCN training is executed on a desktop equipped NVIDIA RTX2070 Super GPU. 
The software environment of all experiments is Python3.9 and CUDA11.2. Please refer to our GitHub link for more details about the environment.
Data normalization, augmentation and early stopping are also used to generate the models as in all most recent state-of-the-art methods. 

\input{Tables/table_comparison_no_ap}

\subsection{Network performance and comparison}

The main result of this work is reported in Table \ref{tab:fusion_all} and Table \ref{tab:comparison}. 
Values are shown as the IoU for the two interest classes, vehicle and human, in different modalities and weather scenarios.
The modalities are indicated as C, L, and C+L, referring to the camera, LiDAR, and fusion, respectively. 

As shown in Table \mbox{\ref{tab:fusion_all}}, the CLFT-hybrid variant outperforms the CLFCN and Panoptic SegFormer in all scenarios, demonstrating high segmentation capabilities over the same data. 
Specifically, in dry environmental conditions, CLFT-hybrid fusion modality archives 91\% IoU for vehicles and 66\% for humans, while CLFCN fusion modality has 90\% for vehicles and 61\% for humans. 
For single modality, Panoptic SegFormer achieves a similar performance of CLFCN for vehicle class but outperforms for human class with less fine-tuned works (61.02\% against 55.57\% in light-dry environment), which reinforces the transformer's strength regarding under-represented samples.
The difference between our CLFT and other models is even more evident in challenging conditions such as dark and wet, where CLFT-hybrid performance drops by 1-2 percentage points while CLFCN and Panoptic SegFormer in single modalities drop by 5-10 percentage points. 
In these cases, fusion seems to play a pivotal role in CLFCN while showing only slight improvements in CLFT-hybrid, demonstrating the robustness of CLFT-hybrid in performing data fusion in all types of conditions.

The Panoptic SegFormer has obvious weak performance in LiDAR modality. 
This is because it is designed to process RGB visual input. 
We carry out the LiDAR processing separately to produce the camera-plane maps with 3D coordinate information; then we feed the maps to Panoptic SegFormer. 
The experiment results prove the necessity to integrate the LiDAR processing into the neural networks' architecture.
Though CLFT-hybrid outperforms the CLFCN in fusion in most cases, it is essential to see that CLFCN models benefit more from the fusion, as the improvement from individual modalities seems to be higher, particularly in night conditions.
On the other hand, our CLFT models already show high performance in challenging conditions with the fusion of camera and LiDAR data.
 
Table \mbox{\ref{tab:comparison}} summarizes the performance of CLFT variants, CLFCN, and Panoptic SegFormer.
We present the precision, recall, and IoU for all models. 
In order to have a straightforward comparison, we combine four weather scenarios for performance evaluation. 
In all cases, the CLFT-hybrid variant performs better than the base and huge variants. 
This result is consistent with what Dosovitskiy et al. \cite{dosovitskiy2020image} reported in their ablation experiments, in which ResNet-based transformer variants outperform the variants that use patch-based embedding procedures.  
Though the CLFT-hybrid achieves the highest IoU score, CLFCN and Panoptic SegFormer have higher recall and precision results, respectively. 

\input{Tables/ablation}

\subsection{Ablation study}

Table \ref{tab:fablation} reports our results using camera (C), LiDAR (L), and fusion (C+L).
According to our ablation study in Table \ref{tab:fablation}, it is possible to conclude that fusion provides an improvement over single-modality networks.  

One might note that results for the individual modalities, particularly LiDAR, show already performance over 90\% (before fusion); this result is also in line with many other studies in the field, for instance, in \cite{zhu2021cylindrical} the authors reached over 90\% IoU in the car class on the SemanticKitti dataset \cite{behley2019iccv}. 

Inspecting the analysis on all-weather, one can see that CLFT-hybrid provides a small improvement (less than one percentage point in both classes).
However, as by construction, the dataset split is strongly unbalanced (see Table \ref{table_datasplit}) toward light-dry scenario (roughly 68\% of the total). 
The amount of light scenarios covers over 88\% of the total number of frames. 
Clearly, the class that is better represented in the dataset affects the overall result the most.

To better appreciate the improvement in our studies, Table \ref{tab:fablation} is also divided according to the data split in Table \ref{table_datasplit}. 
Under these conditions, it is possible to assert that fusion has a higher impact in dark scenarios, covering roughly 12\% of the total number of frames in our dataset.

The unbalance of the dataset has an impact on both environment conditions and object classes, thus the vehicle class (with already over 90\% accuracy) is less affected, while the human class shows better improvements, reaching around 2-4\% in rainy conditions.

\subsection{Inference time analysis}

Table \ref{tab:comp_time} presents an additional study on the inference time. 
In the experiments, we make the statistic of CUDA event time on NVIDIA A100 GPU for fusion modality of all models. All the models are set in evaluation mode for inference time calculation. 
We use the image in Figure \ref{fig: dilated_lidar_anno_imgs} as input, first warm up the GPU with 2000 iterations, then calculate the mean time of the event stream for another 2000 iterations. 
The CPU and GPU are synchronized when recording timestamps. 
In general, FCN-based models have obvious advantages against the transformer-based models in terms of computational efficiency.  
The Panoptic SegFormer has the highest inference time among all models in experiments.
It appears that the CLFCN is faster than our best-performing model, the CLFT-hybrid. 
However, this difference is only about 10ms per frame, which can be considered reasonable in a trade-off between performance and speed. 
For autonomous driving, where safety comes first, classification performance should always be considered a crucial parameter in the network design.
 
\input{Tables/table_compTime}

\subsection{Qualitative results}

\begin{figure*}[th]
\centering
\includegraphics[width=1.0\linewidth]{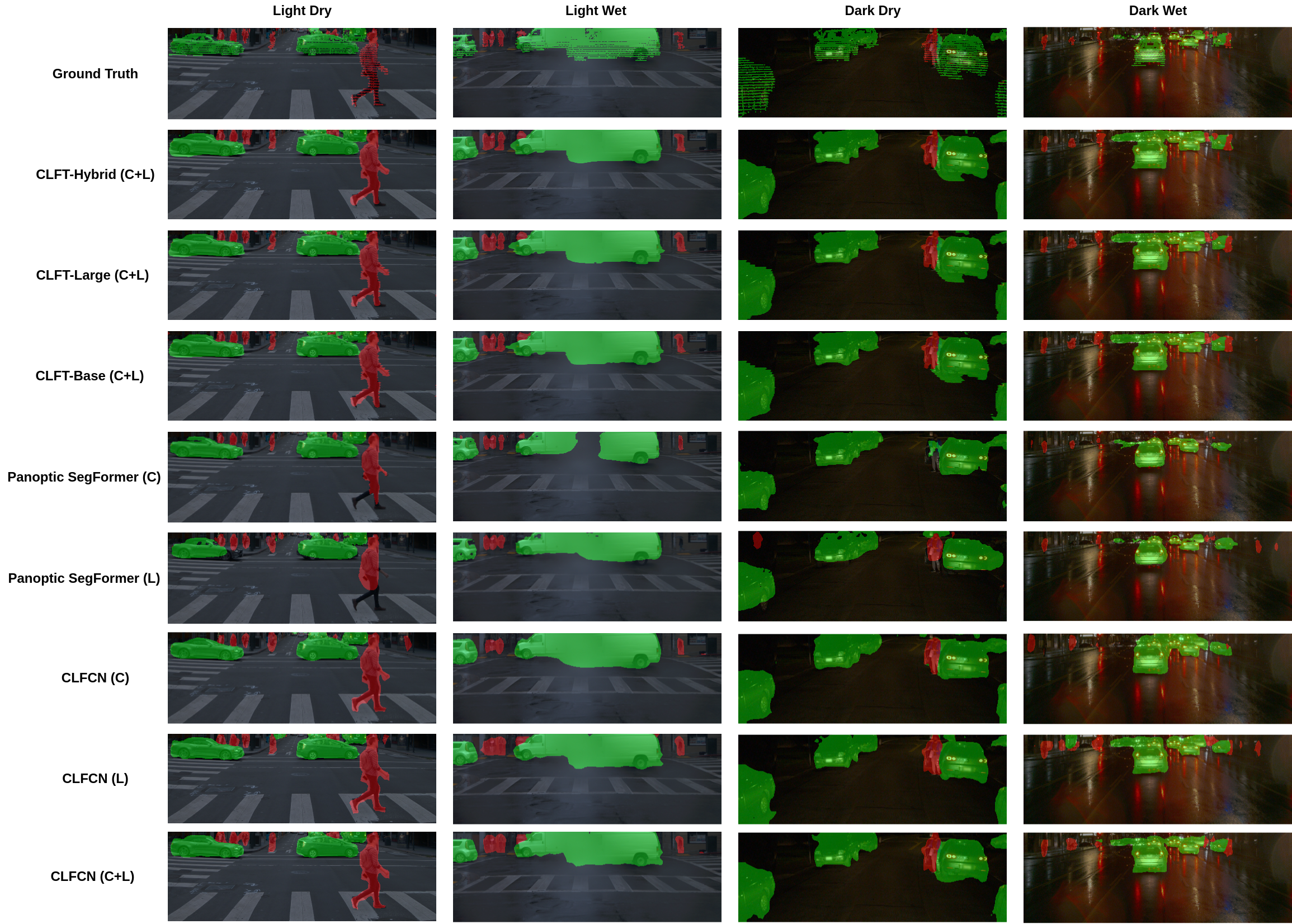}
\caption{Qualitative comparison of segmentation results between different models.}
\label{fig: qualitative_results}
\end{figure*}

Figure \ref{fig: qualitative_results} presents examples of segmented images from the Waymo dataset to appreciate the results of this work from a qualitative point of view. 
Following the above mentioned contribution of this work, the qualitative evaluation is also divided by network structure, weather and illumination conditions. 
The three CLFT variants, `Base', `Large', and `Hybrid', are compared with the Panoptic SegFormer and CLFCN modalities. 
The segmentation results from models are overlaid to the camera images for comparison. 
The first row is the ground truth segmentation provided by the dataset. 
Please note that the annotations of the Waymo dataset are based on the LiDAR point clouds data, which is a common labeling strategy adopted by many famous multi-modal datasets for autonomous driving, including SemanticKitti and nuScenes \cite{caesar2020nuscenes} datasets. 
The LiDAR-points-based labeling strategy results the 2D semantic masks contain the pixels without valid label. 
Waymo dataset claimed to have the highest per-frame point clouds density among the SemanticKitti, nuScenes, and Argoverse \cite{chang2019argoverse} datasets, which is the reason why the Waymo dataset better fits for the evaluation of CLFT networks for 2D semantic segmentation tasks.

The qualitative results generally follow the same consistency as in numerical benchmarks. 
The CLFT-Hybrid variant discloses the most contextual details and its segmentation results are more identical to ground truth than other networks, especially in challenging and under-represented environments. 
For example, the vehicles in night-dry (the third column) scenario, the CLFCN networks detect less details even with fine-tuning efforts, proves that the transformer is more effective than FCN in specific situations. 
Moreover, the single-modality segmentation results from Panoptic SegFormer and CLFCN networks show the necessities and advancements of multi-modal sensor fusion in autonomous driving.

\section{Conclusion}
\label{sec:conclu}
In this paper, we propose a transformer-based multimodal fusion method for semantic segmentation. 
Based on all the above cases, it is possible to say our CLFT model is one of the cutting-edge neural networks for 2D traffic object semantic segmentation. 
Specifically, the CLFT models benefit from the multimodal sensor fusion and transformer's multi-attention mechanism, make a significant improvement for under-represented samples (maximum 10 percent IoU increase for human class). 
However, it is worth mentioning that transformer networks intuitively require a large amount of data for training. 
In our experiments, light-wet and dark-wet subsets only take into account 12\% of the total input data, which explains that the CLFCN model outperforms the CLFT-hybrid model in some cases in Table \ref{tab:fusion_all}.

This work proposes the adoption of a vision transformer's strategy to divide the input image into non-overlapping patches or extract feature patches from CNN feature maps. 
Intuitively, we project and up-sample LiDAR data to dense point clouds images, then design a double-direction network to assemble and cross-fuse the camera and LiDAR representations to achieve final segmentation. 
We maintain the same input dataset splits and configurations in all our experiments and successfully demonstrate the transformer's merit against the FCN regarding object segmentation tasks. 
Specifically, we classify the input data into sub-categories of different illumination and weather conditions dedicated to comprehensively evaluating the models. 
Similar to prior transformer works, we prove its potential on uneven-distributed datasets and under-represented samples. 
At last, we want to highlight that the initiation of CLFT lies on the progress to extend our framework that aims to cover all aspects of low-speed autonomous shuttles, including hardware configuration, dataset collection and post-processing for perception \cite{gu2023endtoend}, validation \cite{malayjerdi2023two}, and path planning \cite{malayjerdi2022practical}. 
We develop the CLFT to be compatible with other systems in terms of environment, data formats, and operating platforms, which grants our work the advantages in scalability and practical application on real autonomous shuttles.

\section*{Acknowledgment}
This research was funded by the European Union's Horizon 2020 Research and Innovation Programme, under the grant agreement No. 856602. This research was co-funded by the European Union under the project Robotics and advanced industrial production (reg. no. CZ.02.01.01/00/22\_008/0004590).

\ifCLASSOPTIONcaptionsoff
  \newpage
\fi



\bibliographystyle{IEEEtran}
\bibliography{bibtex/bib/references}
%



%
\begin{IEEEbiography}[{\includegraphics[width=1in,height=1.25in,clip,keepaspectratio]{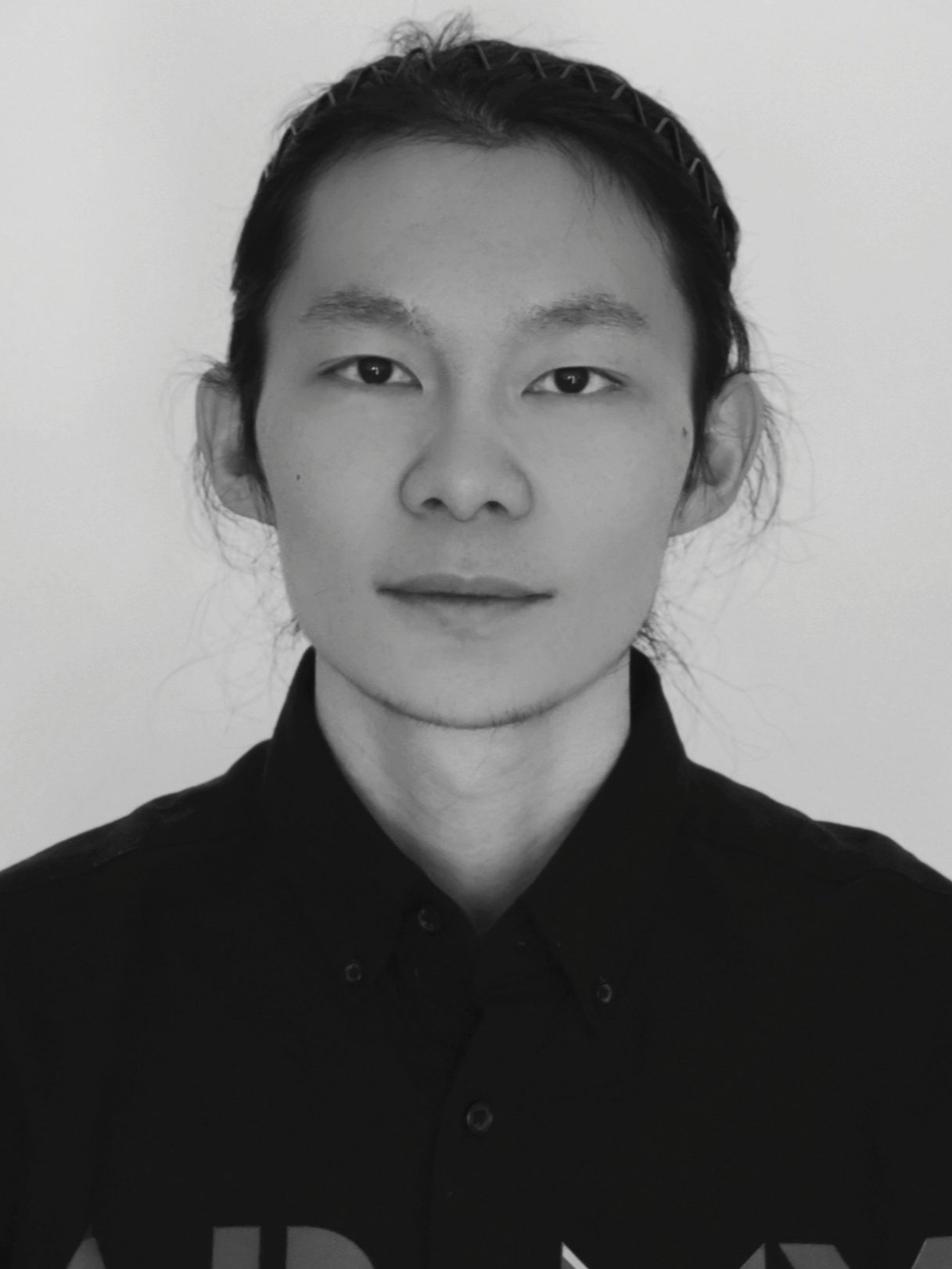}}]{Gu Junyi}
received the B.S. degree in School of
Optical-Electrical and Computer Engineering from the
University of Shanghai for Science and Technology, Shanghai, China, in 2017.
He received the M.S. degree in the Institute of Technology from the University of Tartu, Tartu, Estonia, in 2020.
He is currently pursuing the Ph.D. degree at the Department of Mechanical and Industrial
Engineering, Tallinn University of Technology, Tallinn, Estonia.
His research interests include multi-sensor fusion, semantic segmentation, artificial intelligence, and autonomous driving.
\end{IEEEbiography}

\begin{IEEEbiography}
    [{\includegraphics[width=1in,height=1.25in,clip,keepaspectratio]{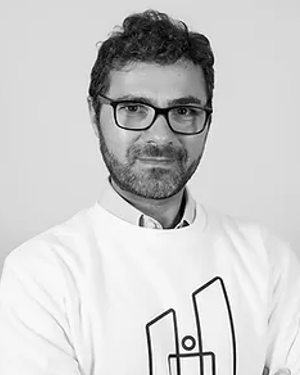}}]{Mauro Bellone}
received his M.S. degree in Automation Engineering from the University of Salento, Lecce, Italy, where he received his Ph.D. in Mechanical and Industrial Engineering in 2014. His interests comprise mobile robotics, autonomous vehicles, energy, computer vision, and control systems. His research focuses on advanced sensory perception for mobile robotics and artificial intelligence. From 2015 to 2020, he worked with the applied artificial intelligence research group of Chalmers University of Technology, where he actively contributed to several autonomous driving projects. In 2021, he was appointed as an adjunct professor at Tallinn University of technology, supporting the research team in the area of smart transportation systems.
\end{IEEEbiography}

\begin{IEEEbiography}
    [{\includegraphics[width=1in,height=1.25in,clip,keepaspectratio]{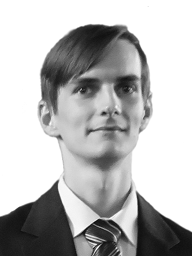}}]{Tom\'a\v{s} Pivo\v{n}ka}
has received his master's degree in robotics at Faculty of Electrical Engineering of Czech Technical University in Prague (CTU) in 2018, where he continues in Ph.D. study program Artificial Intelligence and Biocybernetics. He works at the Intelligent and Mobile Robotics Group of Czech Institute of Informatics, Robotics and Cybernetics, CTU. His main research interests are visual localization, navigation, and computer vision.
\end{IEEEbiography}

\begin{IEEEbiography}
    [{\includegraphics[width=1in,height=1.25in,clip,keepaspectratio]{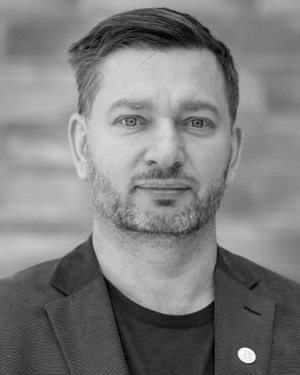}}]{Raivo Sell}
received his Ph.D. degree in Product Development from Tallinn University of Technology in 2007 and currently working as a professor of robotics at TalTech. His research interest covers mobile robotics and self-driving vehicles, smart city, and early design issues of mechatronic system design. He is running the Autonomous Vehicles research group at TalTech as a research group leader with a strong experience and research background in mobile robotics and self-driving vehicles. Raivo Sell has been a visiting researcher at ETH Zürich, Aalto University, and most recently at Florida Polytechnic University in the US, awarded as a Chart Engineer and International Engineering Educator.
\end{IEEEbiography}

\end{document}

%% file: Tables/table_datasplit.tex
\begin{table}[H]
\caption{Amount of the frames in four broad subsets for Waymo Open Dataset.}
\label{table_datasplit}
\centering
\begin{tabular}{|c|c|c|c|}
\hline
Light-Dry & Dark-Dry & Light-Wet & Dark-Wet \\
\hline
14940 & 1640 & 4520 & 900\\
\hline
\end{tabular}
\end{table}

%% file: Tables/table_fusion.tex
\begin{table*}[]
\centering
\caption{Performance comparison of CLFT-hybrid variant, CLFCN and Panoptic SegFormer. Bold indicates the best values in each row per class. (in percentage unit) (C, L, and C+L indicate camera-only, LiDAR-only, and fusion modalities, respectively)}
    \label{tab:fusion_all}
\begin{tabular}{c|cc|cc|cc|cc|cc|cc}
 &
  \multicolumn{2}{c|}{\begin{tabular}[c]{@{}c@{}}CLFT-Hybird\\ (C+L)\end{tabular}} &
  \multicolumn{2}{c|}{\begin{tabular}[c]{@{}c@{}}CLFCN\\  (C)\end{tabular}} &
  \multicolumn{2}{c|}{\begin{tabular}[c]{@{}c@{}}CLFCN\\ (L)\end{tabular}} &
  \multicolumn{2}{c|}{\begin{tabular}[c]{@{}c@{}}CLFCN\\ (C+L)\end{tabular}} &
  \multicolumn{2}{c|}{\begin{tabular}[c]{@{}c@{}}Panoptic SegFormer\\ (C)\end{tabular}} &
  \multicolumn{2}{c}{\begin{tabular}[c]{@{}c@{}}Panoptic SegFormer\\ (L)\end{tabular}} \\
          & Vehicle        & Human          & vehicle & Human & Vehicle & Human & Vehicle        & Human          & Vehicle & Human & Vehicle & Human \\ \hline
Light-Dry & \textbf{91.35} & \textbf{66.04} & 88.08   & 55.57 & 88.58   & 53.04 & 91.07          & 62.50          & 85.89   & 61.02 & 66.41   & 40.78 \\ \hline
Light-Wet & 91.72          & \textbf{66.03} & 88.54   & 52.13 & 89.47   & 50.06 & \textbf{92.77} & 64.66          & 83.58   & 49.70 & 63.07   & 29.87 \\ \hline
Dark-Dry  & \textbf{90.62} & \textbf{65.66} & 81.16   & 42.87 & 86.16   & 48.83 & 89.41          & 60.33          & 81.45   & 44.67 & 70.25   & 38.69 \\ \hline
Dark-Wet  & \textbf{90.18} & 53.51          & 74.49   & 43.14 & 87.51   & 46.68 & 89.90          & \textbf{56.70} & 70.50   & 14.68 & 54.40   & 39.00
\end{tabular}%

\end{table*}

%% file: Tables/table_comparison_no_ap.tex
\begin{table*}[]
\centering
\caption{Performance comparison of all CLFT variants, CLFCN, and Panoptic SegFormer. (in percentage unit)(C, L, and C+L indicate camera-only, LiDAR-only, and fusion modalities, respectively)}
    \label{tab:comparison}
\begin{tabular}{l|ccc|ccc}
 & \multicolumn{3}{c|}{VEHICLE} & \multicolumn{3}{c}{HUMAN} \\ \cline{2-7} 
 & \multicolumn{1}{c|}{Precision} & \multicolumn{1}{c|}{Recall} & IoU & \multicolumn{1}{c|}{Precision} & \multicolumn{1}{c|}{Recall} & IoU \\ \hline
CLFT-Base (C+L) & \multicolumn{1}{c|}{93.63} & \multicolumn{1}{c|}{95.95} & 90.12 & \multicolumn{1}{c|}{71.97} & \multicolumn{1}{c|}{79.47} & 60.68 \\ \hline
CLFT-Large (C+L) & \multicolumn{1}{c|}{93.81} & \multicolumn{1}{c|}{96.14} & 90.46 & \multicolumn{1}{c|}{72.27} & \multicolumn{1}{c|}{77.76} & 60.56 \\ \hline
CLFT-Hybrid (C+L) & \multicolumn{1}{c|}{94.15} & \multicolumn{1}{c|}{96.69} & \textbf{91.26} & \multicolumn{1}{c|}{75.76} & \multicolumn{1}{c|}{82.75} & \textbf{65.46} \\ \hline
CLFCN (C+L) & \multicolumn{1}{c|}{93.17} & \multicolumn{1}{c|}{\textbf{97.67}} & 91.19 & \multicolumn{1}{c|}{65.63} & \multicolumn{1}{c|}{\textbf{92.89}} & 62.51 \\ \hline
Panoptic SegFormer (C)& \multicolumn{1}{c|}{\textbf{94.82}} & \multicolumn{1}{c|}{88.43} &  84.40 & \multicolumn{1}{c|}{\textbf{81.11}} & \multicolumn{1}{c|}{63.78} & 55.55\\ \hline 
Panoptic SegFormer (L)& \multicolumn{1}{c|}{89.57} & \multicolumn{1}{c|}{70.85} &  65.48 & \multicolumn{1}{c|}{67.84} & \multicolumn{1}{c|}{46.85} & 38.29
\end{tabular}
\end{table*}

%% file: Tables/ablation.tex
\begin{table}[ht]
\centering
\caption{Ablation Study based on CLFT-Hybrid variant. (in percentage unit)}(C, L, and C+L indicate camera-only, LiDAR-only, and fusion modalities, respectively)
    \label{tab:fablation}
\centering
\begin{tabular}{cc|cc|cc|cc}
\multirow{2}{*}{C} & \multirow{2}{*}{L} & \multicolumn{2}{c|}{IoU} & \multicolumn{2}{c|}{Precision} & \multicolumn{2}{c}{Recall} \\
   &    & Vehicle & Human & Vehicle & Human & Vehicle & Human \\ \hline \hline
 &    & \multicolumn{6}{c}{All weather}    \\ \hline \hline  
\checkmark &    & 91.16   & 64.38 & 93.86   & 73.33 & 96.88   & 84.05 \\ \hline
   & \checkmark & 91.19   & 65.17 & 93.93   & 72.89 & 96.85   & 84.19 \\ \hline
\checkmark & \checkmark & \textbf{91.26}   & \textbf{65.46} & 94.15   & 75.76 & 96.69   & 82.75 \\ \hline \hline
 &    & \multicolumn{6}{c}{Light-Dry}    \\ \hline \hline
\checkmark &    & 91.23   & 64.87 & 93.83   & 72.63 & 97.05   & 85.86 \\ \hline
   & \checkmark & 91.32   & 64.92 &  93.96  & 72.68 &  97.02  & 85.88 \\ \hline
\checkmark & \checkmark & \textbf{91.35}   & \textbf{66.04} &  94.14  & 75.31 &  96.86  & 84.29 \\ \hline \hline
&    & \multicolumn{6}{c}{Light-Wet}    \\ \hline \hline
\checkmark &    &  91.67  & 64.87 & 94.52   & 76.49 &  96.82  & 81.36 \\ \hline
   & \checkmark &  91.52  & 64.28 &  94.40  & 74.43 &  96.78  & 82.49 \\ \hline
\checkmark & \checkmark &  \textbf{91.72}  & \textbf{66.03} &  94.69  & 78.27 &  96.96  & 80.84\\ \hline \hline
&    & \multicolumn{6}{c}{Dark-Dry}    \\ \hline \hline
\checkmark &    & 90.51   & 65.62 & 93.15  & 74.30 &  96.96  & 84.66 \\ \hline
   & \checkmark & 90.47   & 65.18 & 93.27  & 74.30 &  96.96  & 84.16 \\ \hline
\checkmark & \checkmark &  \textbf{90.62}  & \textbf{65.66}  &  93.38  & 77.39 & 96.68  &81.25\\ \hline \hline
&    & \multicolumn{6}{c}{Dark-Wet}    \\ \hline \hline
\checkmark &    &  89.62  & 52.46 &  93.60  & 70.00 &  95.70  & 67.69 \\ \hline
   & \checkmark &  89.74  & 49.95 &  93.69  & 67.28 &  95.51  & 65.97 \\ \hline
\checkmark & \checkmark &  \textbf{90.18}  & \textbf{53.51} &  94.40  & 68.68 &  95.29  & 70.79
\end{tabular}

\end{table}

%% file: Tables/table_compTime.tex

\begin{table}[h]
\centering
 \caption{Inference time comparison of all CLFT variants, CLFCN and Panoptic SegFormer (in milliseconds unit)(C, L, and C+L indicate camera-only, LiDAR-only, and fusion modalities, respectively)}
     \label{tab:comp_time}
\begin{tabular}{l|c|l}
NETWORK & MODALITY & TIME \\ \hline
CLFT-base                                                 & \multirow{4}{*}{C+L}   & 16.23 \\ \cline{1-1} \cline{3-3} 
CLFT-Large                                                &                        & 36.75 \\ \cline{1-1} \cline{3-3} 
CLFT-Hybrid                                               &                        & 25.69 \\ \cline{1-1} \cline{3-3} 
CLFCN                                                     &                        & 15.94 \\ \hline
\multicolumn{1}{c|}{\multirow{2}{*}{Panoptic SegFormer}}  & C & 93.52 \\ \cline{2-3} 
\multicolumn{1}{c|}{}                                    & L & 93.45 \\ 
\end{tabular}
\end{table}